\documentclass[letterpaper, 10 pt, conference]{IEEEtran}
\IEEEoverridecommandlockouts
\usepackage{algorithm}
\usepackage{algorithmicx}
\usepackage[noend]{algpseudocode}
\usepackage{booktabs} 
\usepackage{multirow} 
\usepackage{amssymb,amsmath,comment}

\usepackage{array}        
\usepackage{makecell}     

\newcommand{\eg}{e.g.,}

\usepackage{xspace}
\usepackage{comment}
\usepackage{listings}
\usepackage{graphicx}
\usepackage{svg}
\svgsetup{inkscapelatex=false}
\usepackage{xcolor}
\usepackage{subcaption}
\usepackage{tikz}
\usetikzlibrary{positioning,arrows.meta,fit,backgrounds,calc,shapes.geometric}
\usepackage{cite}
\usepackage{hyperref}

\usepackage{adjustbox} 
\usepackage{pifont}    

\newif\ifcomments
\commentsfalse
\ifcomments
    \newcommand{\caelan}[1]{{\textcolor[rgb]{0.0,0.0,1.0}{[Caelan: {\it #1}]}}}
    \newcommand{\philip}[1]{{\textcolor[rgb]{0.0,0.1,1.0}{[Philip: {\it #1}]}}}

\else
    \newcommand{\caelan}[1]{}
    \newcommand{\philip}[1]{}
\fi

\begin{document}

\title{\LARGE \bf 
cuArms: GPU-Accelerated Conflict-Based Planning for\\Multi-Arm Manipulation
}


\author{Philip Huang$^{1,2}$, Fabio Ramos$^{1,3}$, Jiaoyang Li$^{2}$, Caelan Garrett$^{*,1}$, Lars Johannsmeier$^{*,1}$
\thanks{$^{*}$ Equal advising, $^{1}$ NVIDIA, $^{2}$Carnegie Mellon, $^{3}$University of Sydney}
}

\maketitle
\DeclareRobustCommand{\methodname}{\textit{cuArms}\xspace}

\begin{abstract}
Multi-arm manipulation has the potential to dramatically improve the throughput, efficiency, and dexterity of many manipulation tasks. 
Recent work coordinates multiple robot arms in a shared workcell by allocating and scheduling tasks for greater parallelism, jointly planning multi-arm motions, or directly learning multi-arm policies.
So far, these approaches are either too slow, rely on domain-specific assumptions, or scale poorly with the number of robots and task horizon.
Thus, we introduce \methodname, a conflict-based planner co-designed with GPU parallelism that efficiently decomposes the search space using reusable multi-modal roadmaps for individual robots and resolves conflicts and deadlocks through a lazy search.
To jointly optimize task assignment, schedules, motions, and manipulation skills, we integrate a high-level scheduler that proposes candidate skeletons for \methodname~to ground into coordinated manipulation plans.
Experimental results show that our method establishes a new state of the art, with real-time initial solutions and anytime improvements in both motion quality and scheduling, without days of domain-specific training required by learning approaches. 
Videos of both simulated and real-world industrial tasks are available on {\urlstyle{same}\nolinkurl{https://cuarms.github.io/}}.
\end{abstract}

\section{Introduction}
As robotic manipulator hardware becomes more widely available and less expensive, multi-arm robotic manipulation---using multiple arms to complete tasks in a shared work cell---becomes much more affordable and accessible. A multi-arm system improves task throughput with greater parallelization, increases space utilization on factory floors, and becomes necessary for tasks that require more dexterous solutions and bimanual coordination, or the handling of bulky parts.

Realizing this promise requires tightly coordinating the motions and manipulation skills of multiple robots that share the same workspace to avoid collisions. 
Specifically, there are several technical challenges: task allocation (which robot is assigned to each task), scheduling (the order and duration of each robot's actions), safety (robots must not collide), and reasoning over hybrid action spaces (e.g., contacts, grasps, kinematic modality changes, reachability).
Current research in this area is often domain-specific~\cite{huang2025apexmr, tian2025fabrica}, time-consuming~\cite{shaoul2024xecbs, hartmann2025benchmark}, 
and cannot plan bimanual manipulation skills~\cite{shaoul2024xecbs, huang2025vampmr}.

\begin{figure}
    \centering
    \includegraphics[width=\linewidth]{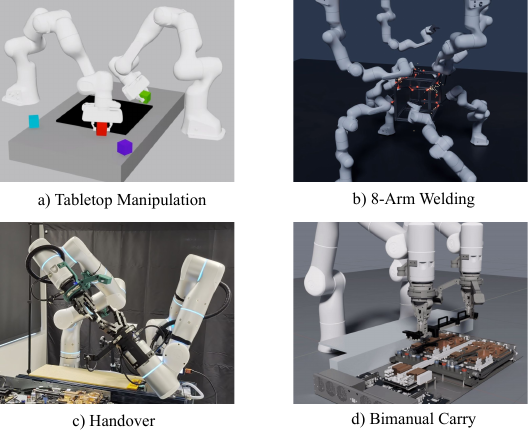}
    \caption{Applications of \methodname: a) tabletop manipulation, b) welding, c) real object reorientation, and d) bimanual carry.
    }
    \label{fig:teaser}
    \vspace{-20pt}
\end{figure}

\begin{figure*}
    \centering
    \includegraphics[width=\linewidth]{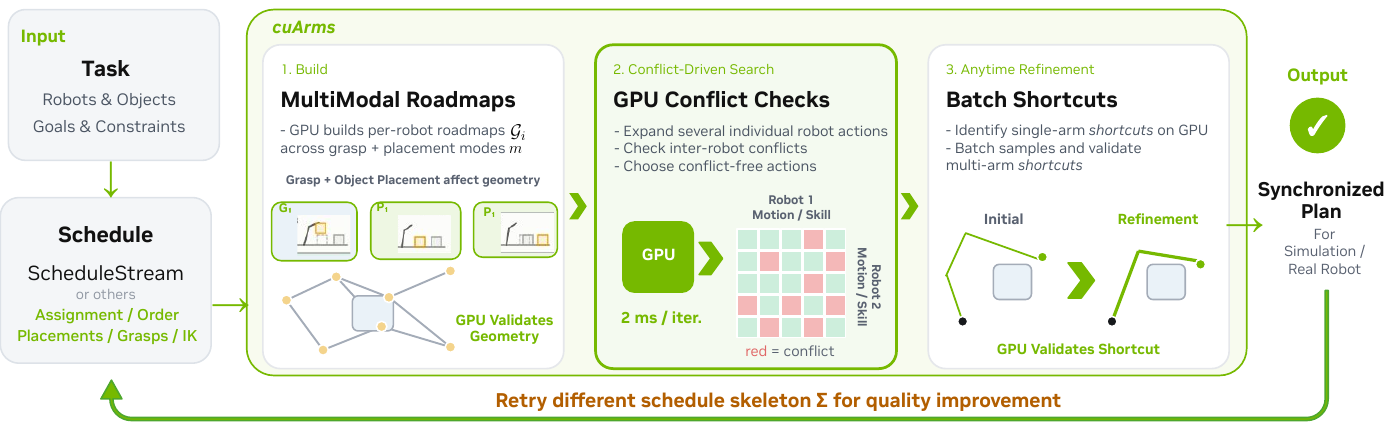}
    \caption{System and \methodname overview:
    A human or agent defines the task: the robots and their kinematics, the objects, goals, and constraints. The domain also supplies samplers that can generate feasible manipulation choices. A task scheduler, such as ScheduleStream, proposes a candidate schedule. This includes task assignment and ordering, along with satisfying samples such as placements, grasps, and inverse kinematic solutions. \methodname~builds a per-robot multi-modal roadmap, searches for conflict-free robot movements and skills with GPU-batched conflict tables, and refines the solution trajectory with shortcutting. \methodname~can iteratively replan with different schedule skeletons to improve upon infeasible or poor skeletons.}
    \label{fig:method}
    \vspace{-10pt}
\end{figure*}

To make these increasingly difficult problems tractable, we introduce \methodname, a new planner that tightly couples GPU acceleration
with a conflict-based multi-agent planner \cite{okumura2023lacam} to find long-horizon, multi-modal, and multi-arm manipulation plans. 
This combination attacks the computational bottleneck from both sides: algorithmic decompositions reduce the search complexity, while GPU parallelism accelerates expensive geometric validations.
\methodname~constructs a multi-modal roadmap~\cite{kavraki1996prm} for each robot, searches the joint configuration space to resolve conflicts among robot motions and manipulation skills, and iteratively improves the resulting trajectory through shortcutting~\cite{geraerts2007creating,huang2025shortcut}. 
A GPU-native implementation accelerates every stage using large batches of parallel geometric validity checks. 
At the same time, our algorithm reuses computation and avoids excessive CPU-GPU communication by decomposing the geometry of different robots, movable objects, and environments.
Simulation and real-robot experiments show that \methodname~finds initial plans in about 1 second across a variety of domains (e.g., Fig.~\ref{fig:teaser}), including an 8-arm, 40-task welding task, and improves solution quality up to $57\%$ with additional computations.
Furthermore, this efficiency makes full multi-arm task and motion planning (TAMP) tractable.
As shown in Fig.~\ref{fig:method}, we use a hierarchical \textit{schedule-then-coordinate} approach that sequentially generates many candidate schedule skeletons and plans the full multi-arm trajectory with \methodname.
\textbf{Our contributions are:}
\begin{itemize}
    \item \methodname, a multi-arm manipulation planner that combines per-arm roadmaps, lazy conflict-driven joint search, and anytime trajectory refinement in a disaggregated architecture co-designed for GPU-batched computation.
    \item Domain-agnostic, multi-arm task and motion planning by integrating \methodname~with a high-level scheduler with support for collaborative bimanual manipulation skills, such as handover and carry.
    \item Simulation and real-robot experiments for multi-arm manipulation, welding, and assembly tasks demonstrating the algorithm's efficiency, quality, and scalability. 
\end{itemize}



\section{Related Work}
Our work builds on prior work in multi-agent pathfinding (MAPF), multi-arm manipulation, and accelerated planning.



\subsection{Multi-Agent Path Finding (MAPF)}
MAPF algorithms~\cite{stern2019mapf} coordinate thousands of robots on discrete space-time graphs to find collision-free paths.
Priority inheritance with backtracking (PIBT)~\cite{okumura2019pibt} is a seminal algorithm that significantly improves speed and scalability in MAPF by resolving conflicts for single-step actions.
However, PIBT is incomplete and can livelock.
Lazy constraints addition search for MAPF (LaCAM)~\cite{okumura2023lacam} embeds PIBT as a lazy expansion step in a complete depth-first search of the joint configuration space. 
db-LaCAM~\cite{moldagalieva2026dblacam} extends LaCAM to kinodynamic multi-robot motion planning with motion primitives. 
However, LaCAM-style search has only been shown for mobile robots, as building graphs and robot collision checking for arms is much more expensive.
\methodname~uses LaCAM-style search for contact-rich multi-arm manipulation tasks with GPU-accelerated search and roadmap construction.

\subsection{Multi-Arm Manipulation Planning}
Recent work has explored how sampling-based motion planning methods~\cite{shome2020drrtstar} or MAPF techniques~\cite{shaoul2024xecbs} can find efficient and collision-free motions for multi-arm systems. 
However, these are limited to single-query start-to-goal motion planning.
Several approaches~\cite{hartmann2025benchmark,shome2019multiarm}
extend sampling-based planners to long-horizon, multi-arm pick-and-place and handover tasks. 
These methods are limited by collision checking as the dominant computational bottleneck.

Task and Motion Planning (TAMP)~\cite{garrett2021integrated} offers a general model-based hybrid planning framework for long-horizon manipulation planning.
ScheduleStream~\cite{garrett2025schedulestream} searches over temporal, durative actions with a code-based domain description to find multi-arm schedule skeletons.
However, ScheduleStream samples conflict-free motions with rejection sampling, which is ineffective if the paths are highly intertwined. 
Other approaches for multi-arm TAMP also depend on postprocessing~\cite{hartmann2023towards} or synchronization~\cite{pan2021general} to plan multi-arm motions. 
In contrast, \methodname~couples individual robot roadmaps with a conflict-driven search in their composite space to find asynchronous trajectories and skills jointly.
Application-driven systems exploit task structure to scale planning for multi-arm assembly~\cite{huang2025apexmr, tian2025fabrica}, but task-specific hierarchies and constraints make them difficult to generalize to new tasks and objectives.
In contrast, \methodname~is domain-agnostic, and a human or AI agent can define new problems
without extending the planner.


\subsection{Accelerated Motion Planning}
One trend to improve motion planning is to accelerate computation with hardware such as CPU SIMD (\eg~VAMP~\cite{thomason2024vamp}) or GPUs (\eg~cuRobo~\cite{sundaralingam2026curobov2}). 
GPU kinematic solvers and planners often directly use optimized libraries such as PyTorch~\cite{sundaralingam2026curobov2} or JAX~\cite{chen2025spasm}, 
making a Python-based codebase possible for better user and agentic experiences. 
A similar trend appears in task and motion planning: cuTAMP~\cite{shen2024cutamp} uses a GPU to evaluate many task skeletons in parallel.
For multiple robot arms, VAMP-MR~\cite{huang2025vampmr} is a recent multi-arm motion planner that vectorizes collision checking with CPU AVX2 instructions without algorithmic modification, but collision checking remains the computational bottleneck.
To reduce this bottleneck using a GPU while avoiding excessive CPU-to-GPU communication, we design \methodname~to evaluate collisions early and in a few large batches rather than naively moving an existing CPU-based planner onto the GPU. 

\section{Multi-Arm Manipulation Planning}

\textit{Robot Workspace.}
We assume the workspace $\mathcal{W}$ contains $n$ robot arms $\mathcal{R}=\{r_1, \dots, r_n\}$, static obstacles, and a set of movable rigid objects $\mathcal{O}$.
Each robot $r_i$ moves in a configuration space $C_i$ subject to its joint position and velocity limits, and has a known base-to-world transform. 
Robots, grippers, and objects in the hand are modeled as unions of collision spheres~\cite{sundaralingam2026curobov2}.

\begin{figure*}
    \centering
    \includegraphics[width=\linewidth]{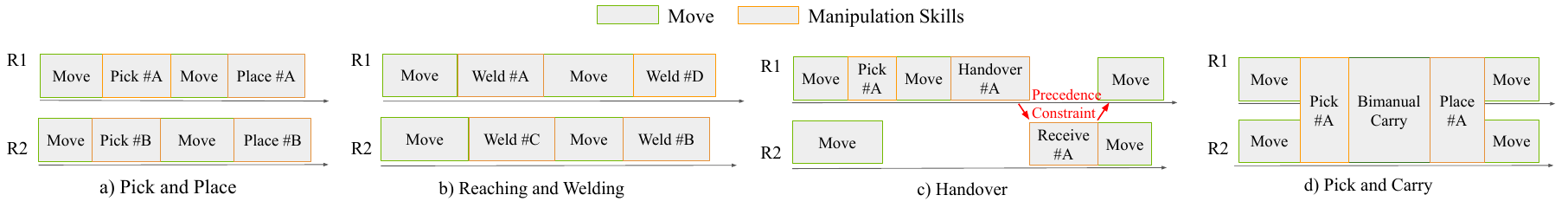}
\caption{Example schedule skeletons in the four domains from Fig.~\ref{fig:teaser}: (a) tabletop pick-and-place, (b) multi-robot reaching for welding, (c) bimanual handover, and (d) bimanual pick and coordinated carry. The high-level scheduler produces the skeleton with corresponding skill parameters such as handover poses, object placements, and grasp poses. \methodname~then computes the coordinated multi-arm motions and skills based on the schedule skeleton.}
    \label{fig:schedules}
\end{figure*}

\textit{Modes.} A \emph{mode} $m\in\mathcal{M}$ specifies the discrete grasp, attachment, or placement state of every object $o\in\mathcal{O}$ relative to a robot frame or the world frame~\cite{hauser2011multimodal}.
\emph{Modes} determine the state of all objects, and combined with the robot configuration fully define the 
state of the system.
Only manipulation skills can change \emph{modes}.
For example, in Fig. \ref{fig:schedules}, robot skills can change the object-robot attachments and thus \emph{modes} via pick and place (Fig. \ref{fig:schedules} (a)) or pick and handover (Fig. \ref{fig:schedules} (c)).
In Fig. \ref{fig:schedules} (d), the robots become kinematically coupled through the shared object during the coordinated pick-carry-place sequence. 

\textit{Skills and Motions.}
Let $\mathcal{A}$ denote the set of manipulation skills. 
Each skill $a_j\in\mathcal{A}$ is a parametrized, contact-rich interaction with one or two arms.
A skill $a_j$ has a set of parameters $x_j = \bigl(q_j^{\mathrm{in}}, q_j^{\mathrm{out}}, d_j, \mathcal{B}_j\bigr)$,
where $q_j^{\mathrm{in}}$ and
$q_j^{\mathrm{out}}$ are its entry and exit joint configurations,
$d_j > 0$ specifies its minimum execution duration,
and $\mathcal{B}_j$ is a bounding volume enclosing the collision geometry of all participating arms and manipulated objects during execution.
Each skill $a_j$ has a known mode-transition function $m^{\mathrm{out}} = T_{a_j}(m^{\mathrm{in}})$ that specifies the exit mode given input mode.
Each robot can also move in free space with or without an attached object.
A \emph{move} is a single-robot, collision-free motion within a fixed mode, specified by one or more target joint configurations provided directly or sampled via inverse kinematics. 

\textit{Schedule Skeleton.}
Let a schedule skeleton 
$
\Sigma
{=}
\left(
\{(\hat{a}_j,\mathcal{R}_j,x_j)\}_{j=1}^{M},
\prec_{\Sigma}
\right)
$
define the high-level symbolic plan and be an input to our multi-arm manipulation planner \methodname.
There are $M$ total manipulation skills and \emph{move} segments among all robots.
Each $\hat{a}_j \in (\mathcal{A} \cup \{\mathrm{move}\})$ is either a skill $a_j$ or a \emph{move} motion with participating robot set $\mathcal{R}_j$ and fixed input parameters $x_j$.
The precedence relation $\prec_{\Sigma}$ imposes a total order for each robot and additional task-level constraints between actions from different robots. 
Skills and movements on different robots that are unordered may execute asynchronously, such as the \emph{move} and \emph{weld} skills in Fig. \ref{fig:schedules} (b). 
The skeleton fixes each skill's mode transition function $T_{a_j}$, and the current global mode is determined by the current skills completed.

Fig. \ref{fig:schedules} shows four skeletons.
In (a), each robot alternates between \emph{move} and single-arm \emph{pick} and \emph{place} skills.
In (b), each arm completes a sequence of \emph{move} and \emph{weld} skills with no cross-robot precedence constraints.
In (c), \emph{handover}~\#A and \emph{receive}~\#A are two single-arm skills coupled by a precedence constraint requiring R1 to hold the object until R2 receives it.
In (d), all manipulation skills are bimanual, so both robots must reach their entry configurations before the bimanual \emph{pick-carry-place} skills and depart from their exit configurations. 

\subsection{Problem Formulation}
\label{sec:problem}
Given a multi-arm workspace $\mathcal{W}$ and a schedule skeleton $\Sigma$, the multi-arm manipulation coordination problem jointly plans
collision-free move trajectories and schedules all actions.
The skeleton fixes the actions, participating robots, skill
parameters, and precedence constraints.
Each \emph{move} trajectory must connect the appropriate skill exit and entry configurations and respect the mode sequence induced by the skills.
A solution is a timed manipulation plan
$
\tau
=
\left(
\{\boldsymbol{\xi}_j\}_{\hat{a}_j=\mathit{move}},\;
\{[t_j,t_j']\}_{j=1}^{M}
\right),
$
where $\boldsymbol{\xi}_j$ is the trajectory of a \emph{move} action
and $[t_j,t_j']$ is the interval assigned to each \emph{move} and skill.
A plan is feasible if it is collision-free and satisfies the ordering and duration constraints of the schedule skeleton. 
At every time $t$, the geometries of moving or waiting robots and the enclosing volumes $\mathcal{B}_j$ of active skills must be collision-free with workspace obstacles and with other robots.
A skill $a_j$ is active during its scheduled interval $[t_j,t_j']$, whose duration must satisfy $t_j'-t_j \geq d_j$.
The schedule must respect precedence constraints: $\hat{a}j \prec_{\Sigma} \hat{a}_k$ implies $t_j' < t_k$.
Robots may wait between actions at their current configurations while maintaining their object attachments and satisfying the same collision constraints.

\textit{Cost.} To support domain-specific preferences such as minimal task completion time, or more Cartesian-like movement, the objective of \methodname~is to minimize
\[
J(\tau)
=
w_{\mathrm{mk}}C_{\mathrm{mk}}(\tau)
+
w_{\mathrm{soc}}C_{\mathrm{soc}}(\tau)
+
w_{\mathrm{sod}}C_{\mathrm{sod}}(\tau),
\]
where $C_{\mathrm{mk}}$, $C_{\mathrm{soc}}$, and $C_{\mathrm{sod}}$ are the makespan, sum of robot completion times, and sum of motion-distance costs.
Each robot's completion time is the time it finishes all assigned \emph{moves} and skills and thereafter remains stationary.
The makespan is the maximum robot completion time.
The distance cost may be measured using joint-space path length, task-space path length, volume-weighted distances traveled by collision spheres, or a weighted combination of these metrics.


\section{cuArms Algorithm}
\begin{figure*}[t]
\centering
\footnotesize
\begin{adjustbox}{valign=t}
\begin{minipage}{0.63\textwidth}
  \centering
  \includegraphics[width=\linewidth]{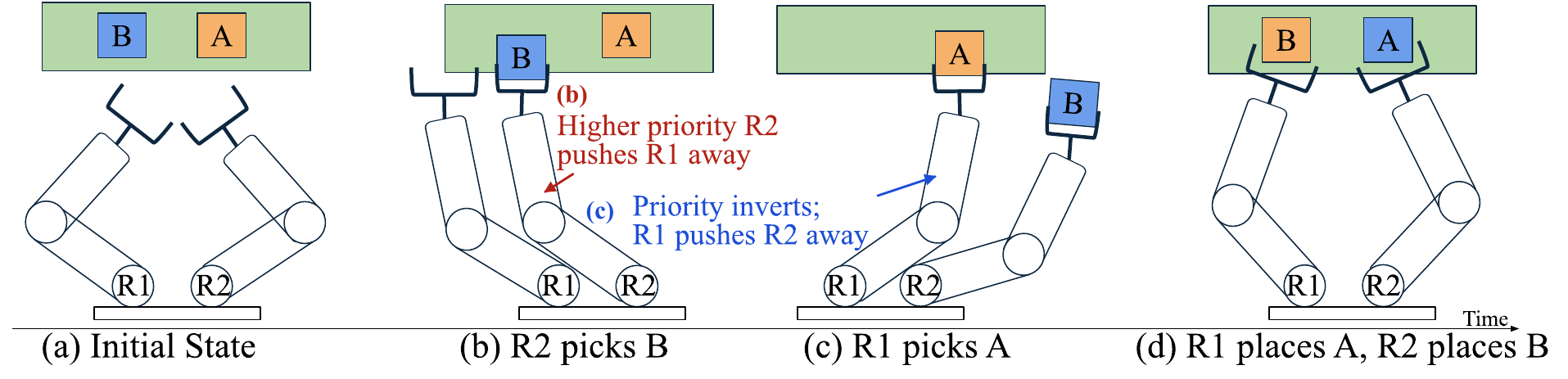}
  \captionof{figure}{A 2D toy object-swapping task. R1 must move A and R2 must move B to exchange their placements. (a) The arms initially block each other. (b) cuArms selects R1 to yield so R2 can pick B. (c) Priority reverses: R2 retreats while holding B so R1 can pick A. (d) Both objects reach their target placements.  }
  \label{fig:pibt-diagram}
\end{minipage}
\end{adjustbox}\hfill
\begin{adjustbox}{valign=t}
\begin{minipage}{0.35\textwidth}
  \centering
  \captionof{table}{Comparison of multi-arm motion or manipulation planning approaches.}
  \label{tab:method-comparison}
  \fontsize{8}{9.5}\selectfont
  \setlength{\tabcolsep}{2pt}
  \renewcommand{\arraystretch}{1.10}
  \begin{tabular}{@{}lcccc@{}}
    \toprule
    Method & Parallelization & Skills & \makecell{Conflict-\\based}   \\
    \midrule
     xECBS~\cite{shaoul2024xecbs} & \ding{55} & \ding{55} & \ding{51}  \\
    Hartmann et al.~\cite{hartmann2025benchmark} & \ding{55} & \ding{51} & \ding{55}   \\
    \addlinespace[2pt]
    APEX-MR~\cite{huang2025apexmr} & \ding{55} & \ding{51} & \ding{55}   \\
    \addlinespace[2pt]
    VAMP-MR~\cite{huang2025vampmr} & \makecell{CPU SIMD} & \ding{55} & \ding{51}   \\
    \addlinespace[2pt]
    ScheduleStream~\cite{garrett2025schedulestream} & GPU & \ding{51} & \ding{55}  \\
    \addlinespace[2pt]
    \midrule
    \textbf{cuArms} & \textbf{GPU} & \ding{51} & \ding{51}  \\
    \bottomrule
  \end{tabular}
\end{minipage}
\end{adjustbox}
\vspace{-10pt}
\end{figure*}

\methodname~disaggregates the multi-arm manipulation coordination problem into three phases: roadmap construction, joint coordination, and anytime refinement. 
The key insight is to separate the reusable single-robot geometry from inter-robot interactions and motion refinement, then co-design each algorithmic phase around GPU-batched geometric computations to maximize efficiency.
Additionally, \methodname~jointly coordinates skills and motions for long-horizon tasks through a conflict-based multi-agent planning method.
This differs from rescheduling solutions that are generated independently~\cite{garrett2025schedulestream} or sequentially~\cite{huang2025apexmr}, or using composite-space sampling and prioritized planning~\cite{hartmann2025benchmark}.
Table~\ref{tab:method-comparison} summarizes these distinctions.
Alg.~\ref{alg:cuarms} provides an overview of our algorithm.

\subsection{Multi-Modal Roadmap Construction}
\label{sec:cuarms-roadmaps}
To simultaneously search over each robot's \emph{move} and skills, \methodname builds a multi-modal roadmap (Alg.~\ref{alg:cuarms}, line~\ref{line:cuarms-roadmap}) with both free-space motions and skill-specific edges.
\methodname~constructs a probabilistic roadmap $\mathcal{G}_i$~\cite{kavraki1996prm} for each robot.
The robot configurations of each node $v_i$ and the connecting edges of $\mathcal{G}_i$ must be collision-free with the static environment, and \methodname~samples and stores valid nodes and edges in GPU memory.
$v_i^\text{init}$ is the initial node.
Roadmaps are well suited to large GPU batches because sampling can be parallelized and cached geometry can be reused across different schedule skeletons.

To account for movable objects $\mathcal{O}$ and modality switches in the skeleton $\Sigma$ without rebuilding the roadmap, \methodname~constructs a mode-validity mask $\mathbf{v}_i^m$ that marks its valid nodes and edges in each possible mode $m$ under the skeleton $\Sigma$.
The GPU batch-checks cached robot configurations against objects at their mode-specific placements, and attached objects against the environment and other objects.
This decomposition reuses forward kinematic calculations and robot--environment collision results across modes, while evaluating the changing robot-object and attached-object interaction in GPU batches without rebuilding roadmaps.
Intuitively, the decomposition works because objects only sparsely impact the validity of a uniformly sampled roadmap, so \methodname~can reuse the same set of sampled nodes and edges for all modes. 

\methodname~then augments the roadmap $\mathcal{G}_i$ with special skill nodes and edges to connect motions and manipulation skills.
For each skill's parameter tuple $x_j$, \methodname~adds the entry $q_j^\text{in}$ and exit $q_j^\text{out}$ configurations as nodes and connects them with an edge representing the skill execution and associated mode transitions.
These entry and exit configurations are also initial and target nodes for neighboring \emph{moves} in the skeleton $\Sigma$. 
The skill's bounding volume $\mathcal{B}_j$ encloses the skill edge for batched collision checking.
Execution durations are rounded up to a whole-number multiple of the max edge interval to synchronize the skill execution with normal roadmap motions. 
Thus, planning skill execution and connecting motions becomes searching for valid edge transitions on $\mathcal{G}_i$. 

To inform the conflict-driven search, \methodname~computes a heuristic cost-to-go $h_i[v, \hat a]$ for each node on the roadmap and current action (Alg.~\ref{alg:cuarms}, line~\ref{line:cuarms-heuristic}). 
Using unit costs for each motion and skill edge, \methodname~runs 
Dijkstra's algorithm backward from each action's target set on the base roadmap without the mode validity mask, then adds an estimate of the motion and skill costs remaining after reaching a target. 
When known precedence constraints from the skeleton $\Sigma$ block a robot's next \emph{move} or skill, \methodname~can instead guide motions to a conflict-free home configuration.
For this purpose, \methodname~ computes a separate cost-to-home heuristic from the home node.

\subsection{GPU-Accelerated Conflict-Driven Search}
\label{sec:cuarms-search}
Given the per-robot roadmaps, \methodname~builds on LaCAM~\cite{okumura2023lacam} to plan multi-arm, multi-task, and multi-modal motion and manipulation skills.
The original LaCAM is a depth-first search algorithm that lazily expands the successor joint state guided by the single-step inter-robot conflict resolution algorithm PIBT~\cite{okumura2019pibt}.
Fig.~\ref{fig:pibt-diagram} illustrates this on a toy example, where R1 must first yield for R2's pick, then R2 retreats with B to make room for R1. 
Planning motion independently cannot resolve this deadlock via retiming alone, while sequential planning must explicitly order the inter-robot actions in the schedule skeleton. 
However, the original LaCAM does not track task-level progress for long-horizon manipulation skills.
Resolving conflicts for arms with PIBT is also slow because the recursive priority inheritance repeatedly invokes expensive inter-robot collision checks.

Compared to LaCAM, \methodname~augments the roadmap node $v_i$ with its current symbolic action $\hat{a}$, and each search node $s$ stores the joint roadmap and action states $\langle v_1,\hat a_1,\dots v_{|R|}, \hat a_{|R|}\rangle$ (line~\ref{line:cuarms-state}).
A constraint queue $s.K$ stores the action-constraint set $\kappa$ that forces certain action selection in PIBT, allowing LaCAM to lazily explore alternative joint successors without enumerating all action combinations upfront (line~\ref{line:cuarms-constraint}).
The joint state determines the current manipulation mode and whether the schedule skeleton allows each robot's next action.
At each state, \methodname~expands a fixed list $\mathcal{U}_i(s)$ of candidate motions, waits, and manipulation skills for each robot (line~\ref{line:cuarms-expand}). 
Skills become available when all participating robots reach their entry nodes and have no blocking precedence constraints.
Completing a skill advances the robots to their exit nodes, advances action stages, and applies any mode transitions. 
\methodname~sorts the candidate list $\mathcal{U}_i(s)$ by first prioritizing manipulation skills, then actions that lead to low cost-to-go at the resulting state $s_i'$ (line~\ref{line:cuarms-rank}).

\textbf{The key innovation that accelerates LaCAM is to compute the geometric conflicts \emph{upfront} and in \emph{parallel} on the GPU (line~\ref{line:cuarms-conflicts}), before the recursive PIBT action selection (line~\ref{line:cuarms-pibt})}.
In the original PIBT, robots move on a shared graph, and each candidate next vertex has at most one current robot occupant.
PIBT processes robots in descending order of dynamic priority so that higher-priority robots choose actions first, and each robot selects its next vertex based on the cost-to-goal heuristic while avoiding higher-priority robots.
If a robot requests a vertex occupied by another robot whose next action is undecided, the occupant temporarily inherits its priority and is recursively asked to move; if it cannot, PIBT backtracks and tries another candidate.
For robot arms, a candidate edge can conflict with several arms simultaneously, and the recursive geometry check to find conflicts with other arms is expensive. 
Running these recursive checks on the GPU requires repeated kernel launches and synchronization overhead, which limits GPU parallelism.
Instead, \methodname~gathers the GPU-resident collision geometry for all candidate edges and batch-checks all inter-robot candidate pairs, using aligned robot trajectories for motions and enclosing volumes for skills. 
It then runs PIBT by looking up the resulting Boolean conflict table during priority inheritance and backtracking.
This design separates the recursive CPU control flow from batched GPU computation and avoids repeated GPU launches and CPU-GPU synchronization.

\methodname~uses PIBT to generate conflict-free joint successors within the lazily constrained LaCAM search.
For bimanual skills, \methodname~commits and constrains participating robots' actions in PIBT and LaCAM together.
\methodname~can also exploit precedence constraints from the skeleton $\Sigma$ if a \emph{move} is blocked by directly prioritizing yield motions with cost-to-home heuristics in candidate ranking.
The search terminates when every robot completes its action sequence, and the solution path is converted into the timed plan $\tau^\star$.

\definecolor{cuarmsgreen}{HTML}{006B45}
\providecommand{\cuarmsnew}[1]{\textcolor{cuarmsgreen}{#1}}

\begin{algorithm}[t]
\footnotesize
\caption{\methodname: Multi-Arm Manipulation Planning}
\label{alg:cuarms}
\begin{algorithmic}[1]
\Require Schedule skeleton $\Sigma$, workspace $\mathcal W$
\Statex \cuarmsnew{$\triangleright$ Green denotes GPU-accelerated steps.}
\ForAll{robots $i\in\mathcal R$}
  \State \cuarmsnew{$\mathcal G_i\gets\Call{MultimodalRoadmap}{i,\Sigma,\mathcal W}$} \label{line:cuarms-roadmap} \Comment{Sec.~\ref{sec:cuarms-roadmaps}}
  \State $h_i\gets\Call{Heuristics}{\mathcal G_i,\Sigma}$ \label{line:cuarms-heuristic}\Comment{Backward Dijkstra}
\EndFor
\State $s^{\text{init}}\gets\langle v_1^{\text{init}},\hat a_1^{\text{init}},\dots v_{|R|}^{\text{init}}, \hat a_{|R|}^{\text{init}}\rangle$
$s^{\text{init}}.K\gets[\varnothing]$; $s^{\text{init}}.\mathrm{parent}\gets\bot$  \label{line:cuarms-state}
\State $S\gets[s^{\text{init}}]$; $\tau\gets\bot$ \Comment{Initialize DFS stack and solution}
\While{$S\neq[\,]$}
  \State $s\gets\Call{Top}{S}$
  \If{all actions in $\Sigma$ are complete at $s$} \Comment{Solution found}
    \State $\tau\gets\Call{Backtrack}{s}$; \textbf{break} \label{line:cuarms-solution}
  \EndIf
  \If{$s.K=[\,]$} \label{line:cuarms-exhausted} \Comment{Constraint queue exhausted}
    \State $\Call{Pop}{S}$; \textbf{continue} \Comment{Skip $s$}
  \EndIf
  \ForAll{robots $i\in\mathcal R$}
    \State $\mathcal U_i(s)\gets\Call{Expand}{s,\mathcal G_i}$ \label{line:cuarms-expand}\Comment{Expand skills, motions, waits}
    \State Sort $\mathcal U_i(s)$ by $\begin{cases}
      (0,h_i[s_i']) & \text{skill},\\
      (1,h_i[s_i']) & \text{motion or wait}.
    \end{cases}$ \label{line:cuarms-rank}
  \EndFor
  \State \cuarmsnew{$C_s\gets\Call{GPUConflictTable}{\{\mathcal U_i(s)\},\{\mathcal G_i\}}$} \label{line:cuarms-conflicts}\Comment{Sec.~\ref{sec:cuarms-search}}
  \State $\kappa\gets\Call{Pop}{s.K}$; enqueue new constraints extending $\kappa$ into $s.K$ \label{line:cuarms-constraint}\label{line:cuarms-enqueue}
  \State $s'\gets\Call{PIBT}{s,\{\mathcal U_i(s)\},C_s,\kappa}$ \label{line:cuarms-pibt}\Comment{Conflict-resolution}
  \If{$s'\neq\bot$ and $s'$ is new}
  \State $s'.K\gets[\varnothing]$; $s'.\mathrm{parent}\gets s$;  $\Call{Push}{S,s'}$\label{line:cuarms-push}
\EndIf

\EndWhile
\If{$\tau=\bot$} \State \Return $\bot$ \EndIf
\State \Return \cuarmsnew{$\Call{BatchedRefinement}{\tau,\Sigma,\mathcal W}$} \Comment{Sec.~\ref{sec:cuarms-refinement}}
\end{algorithmic}
\end{algorithm}

\subsection{Anytime Refinement}
\label{sec:cuarms-refinement}
Because initial solutions found by \methodname~can be suboptimal due to coarse roadmaps and the lack of an optimality guarantee in LaCAM-style search, \methodname~refines solutions by identifying shortcuts for \emph{move} trajectories in batches on the GPU.
Recent work on multi-arm shortcuts~\cite{huang2025shortcut} has proposed how multi-arm shortcuts can improve trajectory quality, but it is nontrivial how shortcuts can improve long-horizon manipulation plans where robots execute both motions and skills asynchronously. 
This is because when post-processing multi-arm plans, optimizing one arm's motion must not introduce collisions with the other robot's motion or skills. 

\methodname~improves its initial plan through batched multi-arm shortcutting co-designed to maximize GPU parallelism.
First, for each \emph{move} trajectory, \methodname~enumerates a bounded set of straight-line interpolations in joint space as shortcuts and batch-evaluates whether these shortcuts are collision-free with the static environment and movable objects $\mathcal{O}$ under its current mode.
From all valid single-robot shortcuts, \methodname~randomly samples a large batch of multi-arm shortcuts.
A multi-arm shortcut can be a single-arm path or prioritized shortcut, where all other arms' trajectories remain the same~\cite{huang2025shortcut} or a composite shortcut made of at least two concurrent single-arm shortcuts. 
\methodname~then evaluates whether these multi-arm shortcuts have any inter-robot conflicts in parallel. 
After finding a set of valid multi-arm shortcuts, \methodname~greedily combines non-overlapping shortcuts in descending order of cost improvement, replacing the original robot trajectory.

\section{Multi-Arm Task and Motion Planning}
\label{sec:tamp}
The full multi-arm TAMP problem requires deciding which robots perform which tasks, in what order, and with which skill parameters, in addition to solving the multi-arm manipulation planning problem.
Joint optimization of the schedule skeleton and coordinated manipulation plan can improve upon using the high-level scheduler or the manipulation planner alone.
A high-level scheduler or temporal planner often solves task allocation and scheduling and evaluates manipulation modes--such as object placements and grasps--effectively, but is less effective at finding tightly coordinated motions.
Optimizing for motions under a fixed schedule skeleton cannot correct poor task assignments or skill parameters that force unnecessary yielding and waiting.

We introduce a hierarchical \textit{schedule-then-coordinate} approach in which a task-level scheduler generates candidate schedule skeletons while \methodname~grounds them into tightly coordinated motion and manipulation trajectories that allow asynchronous execution.
The scheduler may be a general temporal planner such as ScheduleStream~\cite{garrett2025schedulestream}, a domain-specific optimization program, or derived from a fixed-order assembly plan.
If \methodname~fails to find a feasible plan for any candidate skeletons, the system backtracks to the scheduler to generate additional skeletons.
In practice, the scheduler can generate a batch of candidate skeletons upfront, which reduces the need for backtracking.
\methodname also reuses the same multi-modal roadmap stored in GPU memory when the modes are shared; when new modes are introduced, it updates only the mode-validity masks without modifying the base roadmap.
The resulting TAMP hierarchy evaluates many candidate schedule skeletons to improve from an initially infeasible or suboptimal manipulation plan.

\vspace{-3pt}
\section{Experiments}
\vspace{-2pt}

To evaluate \methodname, we examine three questions: 1) How quickly does it find an initial solution? 2) How does it improve solution quality with additional planning time? 3) How does it scale with the number of tasks and robots?

\textit{Environments and Baselines.} We evaluate in two simulated benchmarks as well as real-world bimanual electronic assembly tasks. 
We first evaluate \methodname in a welding task proposed by RoboBallet~\cite{lai2025roboballet}, a state-of-the-art reinforcement learning (RL)-based method.
We retrain the RL policies from scratch in about two days.  
Next, we analyze the performance of \methodname~for TAMP problems in a set of tabletop manipulation tasks against ScheduleStream~\cite{garrett2025schedulestream}, a general temporal planner for multi-arm TAMP and scheduling with GPU acceleration.
Lastly, we show how \methodname~can be deployed in real-world bimanual electronics assembly tasks with integrated manipulation skill sampling and coordinated manipulation planning.


\begin{figure}[t]
    \centering
    \captionsetup[subfigure]{skip=2pt}
    \includegraphics[width=\linewidth]{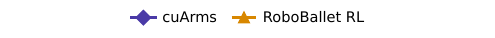}\\[-2pt]
    \begin{subfigure}[t]{0.48\linewidth}
        \includegraphics[width=\linewidth]{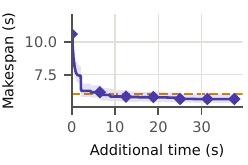}
        \caption{10 targets.}
    \end{subfigure}\hfill
    \begin{subfigure}[t]{0.48\linewidth}
        \includegraphics[width=\linewidth]{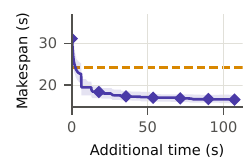}
        \caption{40 targets.}
    \end{subfigure}
    \caption{\textbf{RoboBallet schedule search and refinement.} Execution makespan improves as \methodname~shortcuts initial trajectories and searches alternative schedules.}
    \label{fig:roboballet-anytime}
    \vspace{-10pt}
\end{figure}

\textit{Metrics.} 
For each set of experiments, we evaluate the time to initial solutions $t_0$ and initial and final makespan, denoted by $J_0$ and $J_f$, respectively. 
Unless stated otherwise, each simulation setting uses 15 random seeds; plots show the median and interquartile range (IQR).
We use an NVIDIA RTX Pro 6000 GPU workstation to measure all results.

\subsection{RoboBallet: Many-Arm and Long-Horizon Manipulation}

RoboBallet~\cite{lai2025roboballet} is an RL method for scaling the number of arms and tasks for multi-arm scheduling and motion planning.
The evaluation environment requires 4 or 8 arms performing 10 or 40 spot welding tasks in a shared, tight workspace with obstacles, with the goal of reaching a set of Cartesian weld targets (Fig. \ref{fig:teaser} (b)).
RoboBallet learns a single policy in simulation that jointly allocates and schedules tasks and generates coordinated motions.
RoboBallet provides neither completeness nor formal safety guarantees and also assumes goal-reaching tolerances up to $2.5\,\mathrm{cm}$ and $15^{\circ}$.

Our system combines a greedy scheduler with \methodname~to schedule and plan the full multi-arm reaching task and motions.
The greedy scheduler repeatedly uses Hungarian matching to assign a target to each robot every round using relaxed travel-distance costs until all targets are assigned. 
The joint target configurations for each Cartesian weld target are sampled in batches using GPU-based inverse kinematic solvers~\cite{sundaralingam2026curobov2} to achieve sub-millimeter-level precision.
\methodname~then plans the manipulation trajectory from the given schedule skeleton and refines the initial solutions for 2s and 6s per skeleton for 10- and 40- target problems respectively. 
As only the weld targets are different in each evaluation episode while the workspace remains the same, \methodname~precomputes a base roadmap without any episode-specific joint objectives and reuses it.
During planning, \methodname~first connects each target to the base roadmap using RRT-guided steering. 
Once the system solves the initial skeleton, it backtracks and proposes alternative schedule skeletons with sampled targets for \methodname~to solve iteratively.
A total of 16 schedule skeletons are evaluated.

\begin{table}[t]
    \centering
    \caption{\textbf{RoboBallet scheduling and planning.} $t_\mathrm{offline}$ is the median base roadmap construction time for \methodname and training time for RL. $t_0$ is the total inference time for RL.}
    \label{tab:roboballet-summary}
    \small
    \setlength{\tabcolsep}{2.3pt}
    \begin{tabular}{cc l r rrrr}
        \toprule
        Targets & Robots & Method & Solved & $t_\mathrm{offline} (s)$ & $t_0$ (s) & $J_0$ (s) & $J_f$ (s) \\
        \midrule
        10 & 4 & \methodname & 150/150 & 0.81 & 0.20 & 10.61 & 5.64 \\
        40 & 4 & \methodname & 150/150 & 0.81 & 0.41 & 31.16 & 16.56 \\
        40 & 8 & \methodname & 150/150 & 1.74 & 0.87 & 25.35 & 14.87 \\
        10 & 4 & RL & 29/150 & $1.6\times10^5$ & 0.11 & -- & 6.00 \\
        40 & 4 & RL & 50/150 & $2.2\times10^5$ & 0.42 & -- & 24.25 \\
        \bottomrule
    \end{tabular}
    \vspace{-5pt}
\end{table}

\begin{table}[t]\centering
\caption{\textbf{\methodname median time-to-initial solution (ms) breakdown} in RoboBallet. Roadmap construction refers to connecting new weld targets to the base roadmap. Collision checking includes queries in roadmap construction and the conflict table. }
\label{tab:roboballet-runtime}\small
\setlength{\tabcolsep}{3pt}
\begin{tabular}{@{}lrrr@{}}\toprule
 & \multicolumn{2}{c}{4 arms} & \multicolumn{1}{c}{8 arms} \\
\cmidrule(lr){2-3}\cmidrule(lr){4-4}
Targets & 10 & 40 & 40 \\ \midrule
Greedy Scheduler & 0.59 & 5.12 & 3.06 \\
Roadmap Construction & 98.70 & 155.39 & 271.17 \\
Heuristics & 47.68 & 120.81 & 146.44 \\
Candidate Expansion \& Sort & 8.24 & 24.24 & 32.06 \\
Conflict Table & 22.46 & 65.94 & 277.46 \\
LaCAM search \& PIBT & 6.81 & 13.42 & 19.51 \\
\midrule
Collision Checking (\% of total) & 18.1\% & 24.7\% & 50.2\% \\
\bottomrule\end{tabular}
\vspace{-10pt}
\end{table}

\textit{Results.} \methodname~finds solutions in real time, with a $0.2\,\mathrm{s}$ median time-to-initial-solution for 10 targets and $0.41\,\mathrm{s}$ for 40 targets, each with 10 episodes (Table~\ref{tab:roboballet-summary}).
The base roadmap can be precomputed offline in less than $0.22\,\mathrm{s}$ per robot. 
Shortcutting and evaluating alternative schedules reduce the initial makespan by about $47\%$ in both settings (Fig.~\ref{fig:roboballet-anytime}).
The retrained RoboBallet RL policy completes only $19.3\%$ of 10-target episodes and $33.3\%$ of 40-target episodes, as its predicted actions sometimes fail to avoid collisions and end up in deadlocks, or fail to reach the goal within the $2.5\,\mathrm{cm}$ and $15^{\circ}$ tolerance.
On episodes successfully solved by both methods, \methodname~achieves median makespans that are $20.1\%$ and $27.9\%$ lower than those of RL in the 10-target and 40-target settings, respectively.
When doubling the number of arms to 8, \methodname can reduce the median makespan by $10.2\%$ while still finding feasible solutions within 1 second.
As shown in Table~\ref{tab:roboballet-runtime}, GPU can significantly reduce the collision-checking bottleneck to $18.1$-$24.7\%$ of time-to-initial solution for 4 arms and $50.2\%$ for 8 arms. 
Another major computational bottleneck is constructing the full roadmap that connects the weld targets, which may require more iterative sampling and steering steps towards the base roadmap since the environment obstacles can impose tight corridor-like constraints.


\begin{figure}[t]
    \centering
    \captionsetup[subfigure]{skip=2pt}
    \includegraphics[width=\linewidth]{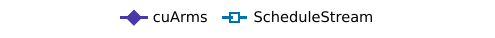}\\[-2pt]
    \begin{subfigure}[t]{0.48\linewidth}
        \includegraphics[width=\linewidth]{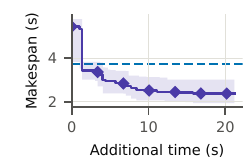}
        \caption{Hold.}
    \end{subfigure}\hfill
    \begin{subfigure}[t]{0.48\linewidth}
        \includegraphics[width=\linewidth]{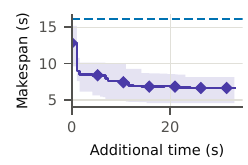}
        \caption{Stack.}
    \end{subfigure}
    \begin{subfigure}[t]{0.48\linewidth}
        \includegraphics[width=\linewidth]{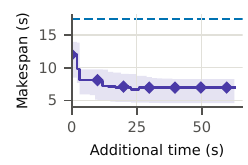}
        \caption{Pack.}
    \end{subfigure}\hfill
    \begin{subfigure}[t]{0.48\linewidth}
        \includegraphics[width=\linewidth]{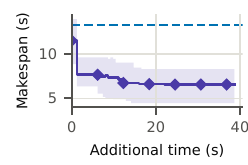}
        \caption{Unpack.}
    \end{subfigure}
    \caption{\textbf{Tabletop manipulation refinement.} \methodname~improves plans across skeletons and manipulation modes. Makespan can increase while the weighted optimization objective improves. }
    \label{fig:schedulestream-anytime}
    \vspace{-4pt}
\end{figure}
\begin{table}[!t]
    \centering
    \caption{\textbf{Multi-arm TAMP for tabletop manipulation} with two to four objects. $J_0$ and $J_f$ are the initial and final makespan, respectively.}
    \label{tab:schedulestream-summary} \small
    \setlength{\tabcolsep}{3pt}
    \begin{tabular}{l rrrr rrr}
        \toprule
        & \multicolumn{4}{c}{\methodname} & \multicolumn{3}{c}{ScheduleStream} \\
        \cmidrule(lr){2-5}\cmidrule(lr){6-8}
        Task & Solved & $t_0$ & $J_0$ & $J_f$ & Solved & $t_0$ & $J_0$ \\
        \midrule
        Hold & 45/45 & 0.48 & 5.46 & 2.37 & 44/45 & 1.75 & 3.74 \\
        Stack & 45/45 & 2.26 & 12.84 & 6.67 & 38/45 & 4.02 & 16.11  \\
        Pack & 45/45 & 2.96 & 12.02 & 6.96 & 38/45 & 4.54 & 17.45  \\
        Unpack & 45/45 & 1.60 & 11.50 & 6.55 & 39/45 & 3.30 & 13.25  \\
        \bottomrule
    \end{tabular}
    \vspace{-10pt}
\end{table}

\begin{figure}[t]
    \centering
    \captionsetup[subfigure]{skip=2pt}
    \includegraphics[width=\linewidth]{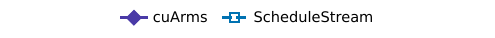}\\[-2pt]
    \begin{subfigure}[t]{0.48\linewidth}
        \includegraphics[width=\linewidth]{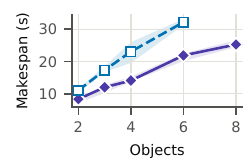}
        \caption{Pack: initial makespan.}
    \end{subfigure}\hfill
    \begin{subfigure}[t]{0.48\linewidth}
        \includegraphics[width=\linewidth]{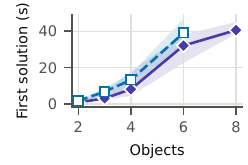}
        \caption{Pack: first-solution time.}
    \end{subfigure}
    \begin{subfigure}[t]{0.48\linewidth}
        \includegraphics[width=\linewidth]{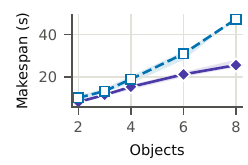}
        \caption{Unpack: initial makespan.}
    \end{subfigure}\hfill
    \begin{subfigure}[t]{0.48\linewidth}
        \includegraphics[width=\linewidth]{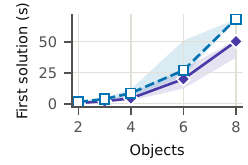}
        \caption{Unpack: first-solution time.}
    \end{subfigure}
    \caption{\textbf{Scaling with number of objects.} \methodname~finds more efficient manipulation plans in less time than ScheduleStream, which also fails to find any solutions for 8-object packing.}
    \vspace{-10pt}
    \label{fig:pack-scaling}
\end{figure}

\begin{figure*}[!t]
    \centering

    \begin{subfigure}[t]{0.495\textwidth}
        \centering
        \includegraphics[width=0.32\linewidth]{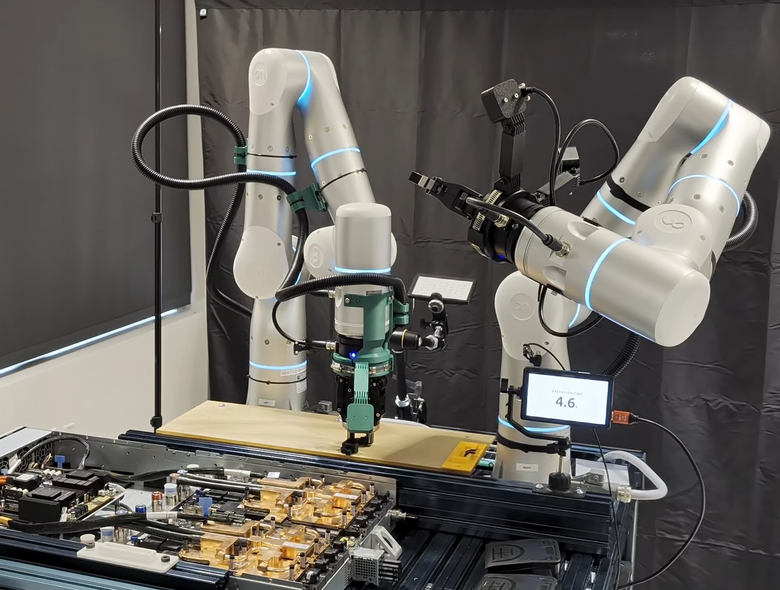}\hfill
        \includegraphics[width=0.32\linewidth]{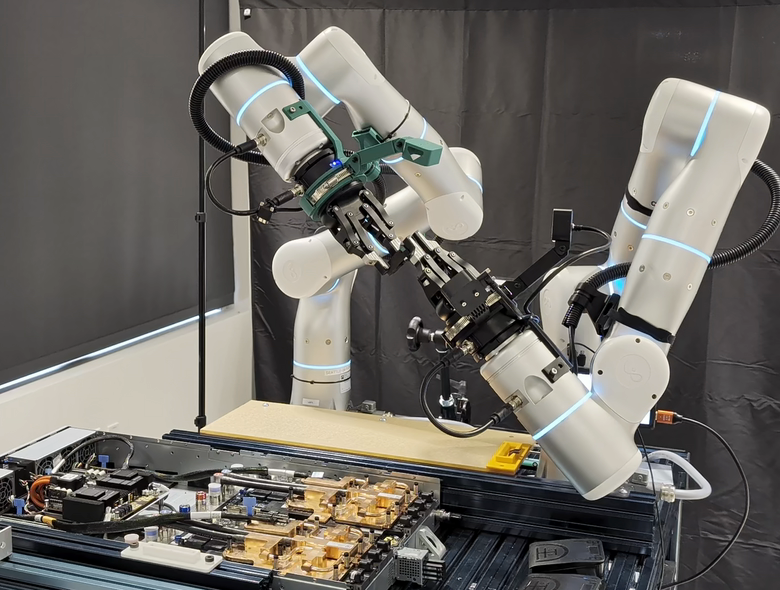}\hfill
        \includegraphics[width=0.32\linewidth]{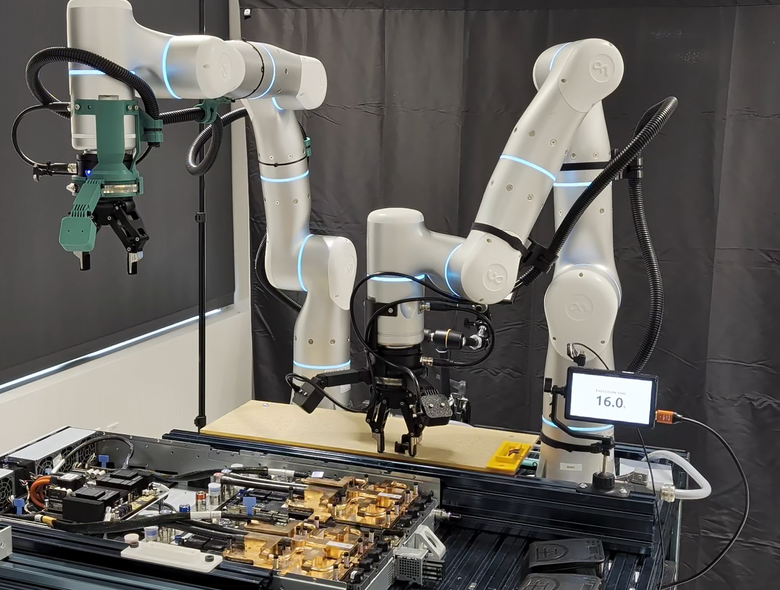}
        \caption{\textbf{Handover.} Re-orient a component in hand using an automatically sampled and optimized handover pose.}
        \label{fig:gb300-handover}
    \end{subfigure}
    \hfill
    \begin{subfigure}[t]{0.495\textwidth}
        \centering
        \includegraphics[width=0.32\linewidth]{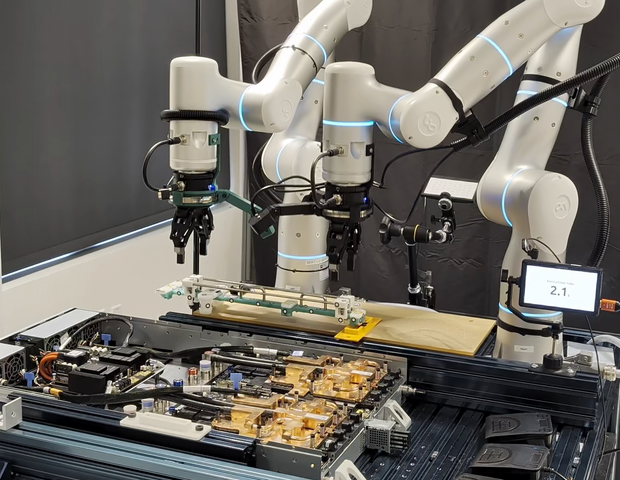}\hfill
        \includegraphics[width=0.32\linewidth]{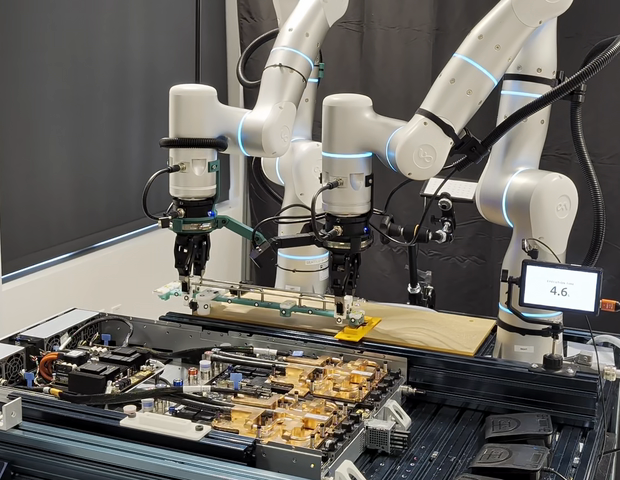}\hfill
        \includegraphics[width=0.32\linewidth]{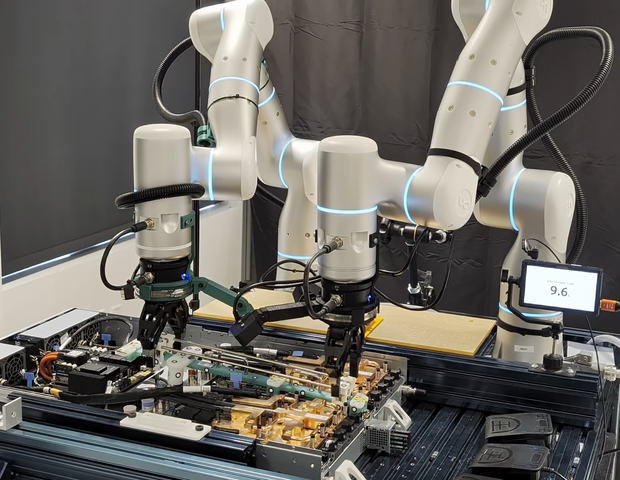}
        \caption{\textbf{Bimanual carry.} Collision-free approach, coordinated grasp, and transport of the busbar fixture into the server tray.}
        \label{fig:gb300-carry}
    \end{subfigure}

    \caption{\textbf{Bimanual Electronic Assembly Tasks with \methodname}.}
    \label{fig:gb300-manipulation}
    \vspace{-15pt}
\end{figure*}

\subsection{Multi-Arm TAMP}
We evaluate our full multi-robot TAMP system for table-top manipulation tasks such as holding, stacking, packing, and unpacking 2-4 cubes with two Panda arms, as shown in Fig.~\ref{fig:teaser} (a).
Packing and unpacking additionally evaluate 6 and 8 objects.
Both \methodname and the ScheduleStream baseline use ScheduleStream's symbolic planner to search over schedule skeletons and sample grasps, placements, and inverse-kinematics solutions with a total high-level search budget of $60\,\mathrm{s}$.
However, the ScheduleStream baseline plans each arm's path independently and then retimes the trajectories to avoid conflicts, whereas \methodname~jointly coordinates the manipulation skills and \emph{move} trajectories.
\methodname~explores up to 16 schedule skeletons with different sampled parameters and refines initial trajectories for $2\,\mathrm{s}$ per skeleton in Pack and $1\,\mathrm{s}$ per skeleton in the other tasks.
\methodname~optimizes the weighted objective with both makespan and path-distance costs. 

\textit{Results.} As shown in Table~\ref{tab:schedulestream-summary}, \methodname~finds initial solutions in $0.48$-$2.96\,\mathrm{s}$, $46.4\%$ faster than ScheduleStream, with a better initial makespan except for the Hold task.
\methodname also solves all instances, while ScheduleStream can fail on some instances requiring tight coordination.
Multi-arm shortcutting and exploration of alternative candidates can efficiently reduce the median \methodname~makespan by $42$--$57\%$ across the four tasks. (Fig.~\ref{fig:schedulestream-anytime}).
Fig.~\ref{fig:pack-scaling} further demonstrates that \methodname can scale up to 8 objects, whereas ScheduleStream finds fewer feasible solutions and higher-makespan plans.

\subsection{Real-Robot Electronics Part Assembly}
Finally, we evaluate the full multi-robot TAMP on two real-world electronics assembly tasks with two Flexiv Rizon4s arms, as shown in Fig.~\ref{fig:gb300-manipulation}. 
The system uses a GPU-accelerated skill sampler to sample constrained poses instead of hardcoded handover poses or carry trajectories and generates schedule skeletons with a ScheduleStream-based symbolic planner. 
\methodname~produces a coordinated manipulation plan with collision-free approach motions around the camera mounts.
Videos of successful execution are available in our supplementary material and {\urlstyle{same}\nolinkurl{https://cuarms.github.io/}}.




\section{Conclusion}


We present \methodname, a new conflict-based multi-arm manipulation planning algorithm.
Using a GPU to batch-evaluate geometry, \methodname~efficiently searches the joint state space with decomposed and reusable per-arm roadmaps and inter-robot conflict resolution. 
\methodname can be integrated into a general multi-arm TAMP algorithm with a \textit{schedule-then-coordinate} approach to improve both low-level coordinated manipulation and high-level scheduling.
With solutions found near-instantaneously without requiring task-specific training, users can use \methodname~to rapidly evaluate the feasibility of multi-arm task and workspace configurations, and integrate it with domain-specific costs and collaborative skills.\footnote{We used OpenAI Codex and Anthropic Claude to assist with code optimization, monitoring experiments, creating plots, and proofreading.}


\bibliographystyle{IEEEtran}
\bibliography{ref}

\end{document}